\documentclass[english]{article}
\usepackage[T1]{fontenc}
\usepackage[latin9]{inputenc}
\usepackage{xcolor}
\usepackage{babel}
\usepackage{array}
\usepackage{multirow}
\usepackage{amstext}
\usepackage{graphicx}
\usepackage[unicode=true]
 {hyperref}

\makeatletter

\providecommand{\tabularnewline}{\\}

\usepackage{spconf}
\usepackage{xcolor}

\usepackage{url}

\name{Anthony Cioppa \qquad Marc Van Droogenbroeck \qquad Marc Braham}
\address{Department of Electrical Engineering and Computer Science (Montefiore Institute)\\ University of Liège, Belgium \\ \{Anthony.Cioppa, M.VanDroogenbroeck, M.Braham\}@uliege.be}

\definecolor{tab_red}{RGB}{231,0,34}
\definecolor{tab_orange}{RGB}{255,114,4}
\definecolor{tab_olive}{RGB}{176,211,41}
\definecolor{tab_green}{RGB}{0,188,50}
\definecolor{tab_blue}{RGB}{35,100,180}

\definecolor{tab_grey}{RGB}{75,75,75}

\usepackage{mathtools}
\usepackage{pifont}

\makeatother

\begin{document}
\global\long\def\SemanticBackgroundSubtraction{\text{SBS}}

\global\long\def\ourMethod{\text{RT-SBS}}

\global\long\def\BackgroundSubtraction{\text{BGS}}

\global\long\def\fscore{F_{1}}

\global\long\def\truepositives{TP}

\global\long\def\truenegatives{TN}

\global\long\def\falsepositives{FP}

\global\long\def\falsenegatives{FN}

\global\long\def\improvement{V^{+}}

\global\long\def\Original#1{I_{#1}}

\global\long\def\B#1{B_{#1}}

\global\long\def\D#1{D_{#1}}

\global\long\def\background{\text{BG}}

\global\long\def\foreground{\text{FG}}

\global\long\def\dontknow{\text{"?"}}

\global\long\def\dontcare{\text{"\ensuremath{\text{\scriptsize{\ding{54}}}}"}}

\global\long\def\ModelSemantic#1{M_{#1}}

\global\long\def\ColorModel{C}

\global\long\def\CtFG{C_{t}^{\foreground}}

\global\long\def\pixel{(x,y)}

\global\long\def\Originalx#1{\Original{#1}\pixel}

\global\long\def\Bx#1{\B{#1}\pixel}

\global\long\def\Dx#1{\D{#1}\pixel}

\global\long\def\px#1{p_{S,#1}\pixel}

\global\long\def\SignalLetter{\text{Si}}

\global\long\def\Signal#1#2{\SignalLetter_{#1}^{#2}\pixel}

\global\long\def\SOutputLetter{S}

\global\long\def\SOutput#1{\SOutputLetter_{#1}\pixel}

\global\long\def\ModelSemanticx#1{\ModelSemantic{#1}\pixel}

\global\long\def\ColorModelx{\ColorModel\pixel}

\global\long\def\Changedetectionx{\ColorModel_{t}\pixel}

\global\long\def\tauBG{\tau_{\background}}

\global\long\def\tauFG{\tau_{\foreground}}

\global\long\def\tauColor{\tau_{C}}

\global\long\def\tauA{\tau_{BG}^{*}}

\global\long\def\tauB{\tau_{FG}^{*}}


\global\long\def\line#1{\text{(L#1)}}


\global\long\def\comma{\,\mbox{,}}

\global\long\def\dot{\,\mbox{.}}


\global\long\def\ruleOne{\text{rule}\,1}

\global\long\def\ruleTwo{\text{rule}\,2}

\global\long\def\ruleA{\text{rule}\,A}

\global\long\def\ruleB{\text{rule}\,B}

\global\long\def\RRuleB{\text{Rule}\,B}

\global\long\def\distanceBetweenTwoValues#1#2{\text{dist}\left(#1,#2\right)}


\global\long\def\ruleMap{R}

\global\long\def\ruleMapx{\ruleMap\pixel}

\global\long\def\colorMap{C}

\global\long\def\colorMapx{\colorMap\pixel}

\global\long\def\inputMap{I}

\global\long\def\inputMapx{\inputMap_{t}\pixel}

\global\long\def\receives{\leftarrow}

\global\long\def\etal{\mathit{et\,al.}}

\global\long\def\etc{\mathit{etc}}

\global\long\def\ie{\mathit{i.e.}}

\global\long\def\eg{\mathit{e.g.}}

\global\long\def\separateur{\mathalpha{:}}

\global\long\def\samplingFactor#1{#1\separateur1}

\global\long\def\framePerSecond{\text{fps}}


\global\long\def\inputFrame{I}

\global\long\def\semanticFrame{S}

\global\long\def\backgroundSegmentedFrame{B}

\global\long\def\postProcessedFrame{D}

\global\long\def\outputFrame{O}

\global\long\def\timeIntervalBetweenFrames{\text{\ensuremath{\delta}}}

\global\long\def\timeCalculation{\Delta}

\global\long\def\timeCalculationFeedback{\timeCalculation_{F}}

\global\long\def\fps{\text{fps}}

\global\long\def\ms{\text{ms}}

\title{Real-Time Semantic Background Subtraction}
\maketitle
\begin{abstract}
Semantic background subtraction ($\SemanticBackgroundSubtraction$)
has been shown to improve the performance of most background subtraction
algorithms by combining them with semantic information, derived from
a semantic segmentation network. However, $\SemanticBackgroundSubtraction$
requires high-quality semantic segmentation masks for all frames,
which are slow to compute. In addition, most state-of-the-art background
subtraction algorithms are not real-time, which makes them unsuitable
for real-world applications. In this paper, we present a novel background
subtraction algorithm called Real-Time Semantic Background Subtraction
(denoted $\ourMethod$) which extends $\SemanticBackgroundSubtraction$
for real-time constrained applications while keeping similar performances.
$\ourMethod$ effectively combines a real-time background subtraction
algorithm with high-quality semantic information which can be provided
at a slower pace, independently for each pixel. We show that $\ourMethod$
coupled with ViBe sets a new state of the art for real-time background
subtraction algorithms and even competes with the non real-time state-of-the-art
ones. Note that we provide python CPU and GPU implementations of $\ourMethod$
at \href{https://github.com/cioppaanthony/rt-sbs}{https://github.com/cioppaanthony/rt-sbs}.

\end{abstract}

\section{Introduction\label{sec:Introduction}}

Background subtraction ($\BackgroundSubtraction$) algorithms aim
at detecting pixels belonging to moving objects in video sequences~\cite{Bouwmans2014Traditional}.
Generally, a $\BackgroundSubtraction$ algorithm is composed of three
elements: an adaptive background model, a similarity criterion to
compare a pixel of a frame with the model, and an update strategy
for the background model. The $\BackgroundSubtraction$ algorithm
then classifies each pixel of the video into one of the following
two classes: foreground ($\foreground$) for moving objects, or background
($\background$). 

While many progresses in background subtraction have been achieved
since the seminal algorithms GMM~\cite{Stauffer1999Adaptive} and
KDE~\cite{Elgammal2000NonParametric}, partly due to the availability
of pixel-wise annotated datasets such as BMC~\cite{Vacavant2012ABenchmark},
CDNet 2014~\cite{Wang2014AnExpanded} or LASIESTA~\cite{Cuevas2016Labeled},
modern algorithms such as ViBe~\cite{Barnich2011ViBe}, PAWCS~\cite{StCharles2016Universal},
or IUTIS-5~\cite{Bianco2017Combination} remain sensitive to dynamic
backgrounds, illumination changes, shadows, $\etc$. Furthermore,
most state-of-the-art algorithms are unusable in practice since they
are not real-time as stated in~\cite{Roy2018RealTime}. More recently,
deep learning based algorithms emerged with the work of\textit{\emph{
}}Braham $\etal$\textit{\emph{~\cite{Braham2016Deep} which opened
the path for novel algorithms~\cite{Bouwmans2020Background,Zheng2018Background-chinese}
thanks to the increased power of computers. }}
\begin{figure}[t]
\begin{centering}
\includegraphics[width=1\columnwidth]{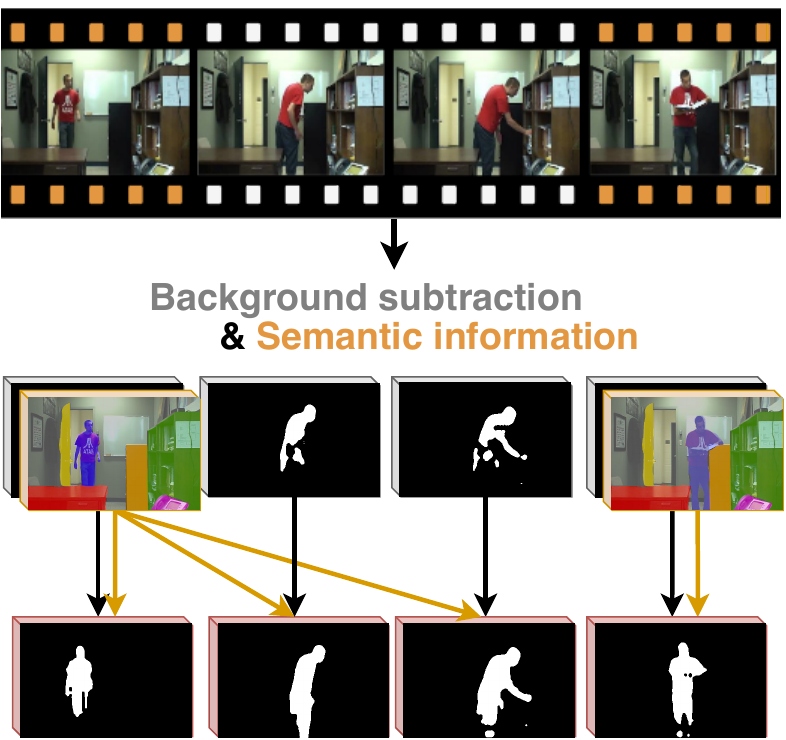}
\par\end{centering}
\caption{Our novel background subtraction algorithm, called ${\color{red}\protect\ourMethod},$
combines a \textcolor{tab_grey}{$\protect\BackgroundSubtraction$
algorithm} with \textcolor{orange}{semantic information} in real
time. Semantic information is slow to compute and is only available
for some frames, but $\protect\ourMethod$ reuses previous semantic
information when appropriate.\label{fig:graphical}}
\end{figure}

In this paper, we focus on a particular $\BackgroundSubtraction$
algorithm developed by Braham $\etal$~\cite{Braham2017Semantic},
called semantic background subtraction ($\SemanticBackgroundSubtraction$).
This algorithm combines the result of a network such as PSPNet~\cite{Zhao2017Pyramid},
used to provide semantic information about objects of interest in
the scene, with a $\BackgroundSubtraction$ algorithm in order to
improve the performance of the latter. The objective of semantic
segmentation consists in labeling each pixel of an image with a class
corresponding to the object that the pixel belongs to. $\SemanticBackgroundSubtraction$
uses semantic information derived from semantic segmentation to infer
whether or not a pixel belongs to a potentially moving object. It
has been shown to improve the performance of most unsupervised and,
more recently, even supervised $\BackgroundSubtraction$ algorithms~\cite{Tezcan2020BSUVNet}.
However, the challenge for using $\SemanticBackgroundSubtraction$
is that producing high-quality semantic segmentation is time consuming
and that the best semantic networks do not process video frames in
real time~\cite{Minaee2020Image}.

In this work, we propose a novel algorithm, called Real-Time Semantic
Background Subtraction ($\ourMethod$) and illustrated in Figure~\ref{fig:graphical},
which is capable to use semantic information provided at a slower
pace, and for some pixels. It reuses previous semantic information
by integrating a change detection algorithm during the decision process.
The latter checks if the last decision enforced by the semantic information
is still up to date to be replicated. This allows our algorithm to
keep performances close to the ones of $\SemanticBackgroundSubtraction$,
while being real time. Furthermore, we introduce a semantic feedback
to further improve the performance of $\SemanticBackgroundSubtraction$.

The paper is organized as follows. Section~2 presents $\SemanticBackgroundSubtraction$
under a novel point of view which then allows to introduce $\ourMethod$.
Then, in Section~3, we show state-of-the-art performance of our
algorithm on the CDNet 2014 dataset and compare it to several other
algorithms. Finally, we conclude the paper in Section~4. \newline\textbf{Contributions.}
We summarize our contributions as follows. \textbf{(i)} We present
a novel real-time background subtraction algorithm that leverages
high-quality semantic information which can be provided at any pace
and independently for each pixel. \textbf{(ii)} We show state-of-the
art performance on the CDNet 2014 dataset with our real-time algorithm.

\section{Real-time semantic background subtraction\label{sec:Real-time-semantic-background}}

Recently, Braham~$\etal$ have introduced the Semantic Background
Subtraction ($\SemanticBackgroundSubtraction$) algorithm~\cite{Braham2017Semantic}
that leverages semantic information to help addressing the challenges
of background subtraction. One major constraint with $\SemanticBackgroundSubtraction$
is that it requires reliable semantic information and that the best
current networks are far from a real-time frame rate. Hence, $\SemanticBackgroundSubtraction$
cannot be used in real-time constrained applications. In this work,
we propose a novel background subtraction algorithm that extends $\SemanticBackgroundSubtraction$
for real-time applications, regardless of the semantic network processing
speed.

\subsection{Description of semantic background subtraction}

$\SemanticBackgroundSubtraction$ combines the decision of two classifiers
that operate at each pixel $\pixel$ and for each frame (indexed by
$t$): (1) a background subtraction algorithm, which is a binary classifier
between the background ($\background$) and the foreground ($\foreground$),
whose output is denoted by $\Bx t\in\{\background,\foreground\}$,
and (2) a semantic three-class classifier, whose output is denoted
by $\SOutput t\in\{\background,\foreground,\text{\ensuremath{\dontknow}}\}$,
where the third class, called the ``don't know'' ($\dontknow$)
class, corresponds to cases where the semantic classifier is not able
to take a decision. This semantic classifier is built upon two signals
that contribute to take a decision. The first one is the semantic
probability that pixel $\pixel$ belongs to a set of objects most
likely in motion. If this signal is lower than some threshold $\tauBG$,
$\SOutput t$ is set to $\background$. The second signal is a pixelwise
increment of semantic probability for a pixel $\pixel$ to belong
to a moving object. When this signal is larger than another threshold
$\tauFG$, $\SOutput t$ is set to $\foreground$. In all other cases,
$\SOutput t$ is undetermined and is assigned a ``don't know'' class,
denoted $\dontknow$ in the following.

Finally, the output of $\SemanticBackgroundSubtraction$, noted $\Dx t$,
is a combination of $\Bx t$ and $\SOutput t$, as outlined in Table~\ref{tab:decision-table-sbs}.
This combination works as follows: when $\SOutput t$ is determined
(either $\background$ or $\foreground$), this class is chosen as
the output of $\SemanticBackgroundSubtraction$ regardless of the
value of $\Bx t$; when $\SOutput t$ is undetermined (which corresponds
to $\dontknow$ cases), the class of $\Bx t$ is chosen as $\Dx t$.
\begin{table}[t]
\begin{centering}
{\footnotesize{}}%
\begin{tabular}{|c|c|c||c|}
\cline{2-4} 
\multicolumn{1}{c|}{} & \multicolumn{3}{c|}{\emph{Decision table of $\SemanticBackgroundSubtraction$}}\tabularnewline
\cline{2-4} 
\multicolumn{1}{c|}{} & \multicolumn{2}{c||}{\textbf{Classifiers}} & \textbf{Output}\tabularnewline
\cline{2-4} 
\multicolumn{1}{c|}{} & $\B t\pixel$ & $\SOutput t$ & $\D t\pixel$\tabularnewline
\hline 
$\line 1$ & $\background$ & $\dontknow$ & $\background$\tabularnewline
\hline 
$\line 2$ & $\background$ & $\background$ & $\background$\tabularnewline
\hline 
$\line 3$ & $\background$ & $\foreground$ & $\foreground$\tabularnewline
\hline 
$\line 4$ & $\foreground$ & $\dontknow$ & $\foreground$\tabularnewline
\hline 
$\line 5$ & $\foreground$ & $\background$ & $\background$\tabularnewline
\hline 
$\line 6$ & $\foreground$ & $\foreground$ & $\foreground$\tabularnewline
\hline 
\end{tabular}{\footnotesize\par}
\par\end{centering}
\centering{}\caption{Decision table for the output of $\protect\SemanticBackgroundSubtraction$
($\protect\Dx t$) based on the output of two classifiers: a $\protect\BackgroundSubtraction$
algorithm ($\protect\Bx t$) and a semantic classifier ($\protect\SOutput t$).
\label{tab:decision-table-sbs}}
\end{table}

While $\SemanticBackgroundSubtraction$ is effective to handle challenging
$\BackgroundSubtraction$ scenarios, it can only be real time if both
classifiers are real time. As the decision of the semantic classifier
supersedes that of $\B t\pixel$ in two scenarios (see lines $\line 3$
and $\line 5$ in Table~\ref{tab:decision-table-sbs}), it is essential
to rely on a high-quality semantic segmentation, which is not achievable
with faster semantic networks.

Another way to reduce the computation time of semantic information
is to segment small portions of the image or to skip some frames.
However, according to the original decision table of $\SemanticBackgroundSubtraction$,
this would introduce more ``don't know'' cases for the semantic
classifier (equivalent to lines $\line 1$ and $\line 4$ of Table~\ref{tab:decision-table-sbs}).
Our algorithm aims at providing a decision different from the ``don't
know'' case when the semantic segmentation has not been calculated
for pixel $\pixel$ at time $t$.

\subsection{Change detection for replacing missing semantic information}

We propose a novel algorithm that reuses previous decisions of the
semantic classifier in the absence of semantic information. We choose
to rely on previously available semantic information and we check
whether or not this information is still relevant. If the pixel has
not changed too much, its predicted semantic class is still likely
untouched, and therefore the previous decision of the semantic classifier
is replicated.

Technically, we introduce a third classifier in the previous decision
table. This classifier corresponds to a change detection algorithm
whose task is to predict whether or not a pixel's value has significantly
changed between the current image, at time $t$, and the last time
semantic information was available for that pixel, at time $t^{*}\leq t$.
The new decision table is presented in Table~\ref{tab:decision-table-rtsbs}
and works as follows: if the change detection algorithm, whose output
is denoted by $\Changedetectionx$, predicts that the pixel has not
changed, it means that the pixel still probably belongs to the same
object and thus the previous semantic decision is repeated. Alternatively,
when the change detection algorithm predicts that the pixel has changed
too much, the previous semantic information cannot be trusted anymore,
leaving it to the $\BackgroundSubtraction$ algorithm to classify
the pixel. The improvement of our algorithm compared to $\SemanticBackgroundSubtraction$
originates from lines $\line 3$ and $\line 6$ of Table~\ref{tab:decision-table-rtsbs},
as without the change detection classifier, the final decision would
be taken by the $\BackgroundSubtraction$ algorithm alone. 
\begin{table}[t]
\begin{centering}
{\footnotesize{}}%
\begin{tabular}{|c|c|c|c||c|}
\cline{2-5} 
\multicolumn{1}{c|}{} & \multicolumn{4}{c|}{\emph{Decision table of $\ourMethod$}}\tabularnewline
\cline{2-5} 
\multicolumn{1}{c|}{} & \multicolumn{3}{c|}{\textbf{Classifiers}} & \textbf{Output}\tabularnewline
\cline{2-5} 
\multicolumn{1}{c|}{} & $\B t\pixel$ & $\SOutput{t^{*}}$ & $\Changedetectionx$ & $\D t\pixel$\tabularnewline
\hline 
$\line 1$ & $\background$ & $\dontknow$ & $\dontcare$ & $\background$\tabularnewline
\hline 
$\line 2$ & $\background$ & $\background$ & $\dontcare$ & $\background$\tabularnewline
\hline 
$\line 3$ & $\background$ & $\foreground$ & No Change & $\foreground$\tabularnewline
\hline 
$\line 4$ & $\background$ & $\foreground$ & Change & $\background$\tabularnewline
\hline 
$\line 5$ & $\foreground$ & $\dontknow$ & $\dontcare$ & $\foreground$\tabularnewline
\hline 
$\line 6$ & $\foreground$ & $\background$ & No Change & $\background$\tabularnewline
\hline 
$\line 7$ & $\foreground$ & $\background$ & Change & $\foreground$\tabularnewline
\hline 
$\line 8$ & $\foreground$ & $\foreground$ & $\dontcare$ & $\foreground$\tabularnewline
\hline 
\end{tabular}{\footnotesize\par}
\par\end{centering}
\centering{}\caption{Decision table of $\protect\ourMethod$. Its output ($\protect\Dx t$)
depends on three classifiers: a $\protect\BackgroundSubtraction$
algorithm ($\protect\Bx t$), information about the last time, $t^{*}\leq t$,
the semantic classifier ($\protect\SOutput{t^{*}}$) classified the
pixel, and a change detection algorithm ($\protect\Changedetectionx$).
The ``don't care'' values ($\protect\dontcare$) represent cases
where $\protect\Changedetectionx$ has not impact on $\protect\Dx t$,
either because previous semantic information is undetermined or because
$\protect\Bx t$ and $\protect\SOutput{t^{*}}$ agree on the class.\label{tab:decision-table-rtsbs}}
\end{table}

The only requirement for the choice of the change detection algorithm
is that it has to be real time. In $\ourMethod$, we choose a simple
yet effective algorithm that relies on the Manhattan distance between
the current pixel's color value and its previous color value when
semantic information was last available, at time $t^{*}$. If this
color distance is smaller (resp. larger) than some threshold, the
change detection algorithm predicts that the pixel has not changed
(resp. has changed). Let us note that we use two different thresholds
depending on the output of the semantic classifier ($\tauA$ if $\SOutput{t^{*}}=\background$,
or $\tauB$ if $\SOutput{t^{*}}=\foreground$) since the foreground
objects and the background change at different rates. In the case
where semantic information is available, the change detection algorithm
will obviously always predict that the pixel has not changed since
$t^{*}=t$, and the decision table of $\ourMethod$ (Table~\ref{tab:decision-table-rtsbs})
degenerates into that of $\SemanticBackgroundSubtraction$ (Table~\ref{tab:decision-table-sbs}).

\subsection{Introducing a semantic feedback}

The last choice to make is the one of a real-time $\BackgroundSubtraction$
algorithm. We choose to use ViBe~\cite{Barnich2009ViBe} as it is
the best real-time $\BackgroundSubtraction$ algorithm according to~\cite{Roy2018RealTime}.
This algorithm has the particularity of updating its background model
in a conservative way, meaning that only pixels classified as background
(or close to a background pixel) can be updated. Instead of keeping
the output of ViBe for the update ($\Bx t),$ we replace it with the
output of $\ourMethod$ ($\Dx t$) which is better. This introduces
a feedback of the semantic information in the background model (via
lines $\line 3$ and $\line 6$ of Table~\ref{tab:decision-table-rtsbs}),
which makes ViBe take better decisions at each new frame. It is interesting
to note that $\ourMethod$ could be used with any other $\BackgroundSubtraction$
algorithm, just like $\SemanticBackgroundSubtraction$, and that the
semantic feedback is an add-on in the case of ViBe.

\section{Experimental results\label{sec:Experimental-results}}

For the evaluation, we choose the CDNet 2014 dataset~\cite{Wang2014AnExpanded}.
It is composed of $53$ video sequences mostly shot at $25$ fps and
comprising some challenging scenarios such as intermittent object
motion, dynamic backgrounds and moving cameras. We use the overall
$\fscore$ score as suggested in the evaluation policy to compare
$\ourMethod$ with the other $\BackgroundSubtraction$ algorithms.
Semantic segmentation is computed as in~\cite{Braham2017Semantic}
using the semantic segmentation network PSPNet~\cite{Zhao2017Pyramid}.
We consider that semantic information is available for one in every
$X$ frame, which is denoted as $\samplingFactor X$. As PSPNet runs
at about $5$ frames per second (fps) on the images of the dataset
according to~\cite{Braham2017Semantic}, only a maximum of $1$ out
of $5$ frames can have access to the semantic information in a real-time
setup, which is denoted as $\samplingFactor 5$. In our experiments,
the performance is computed for several semantic frame rates $\samplingFactor X$.
Real-time configurations correspond to $X\geq5$.

The parameters of $\ourMethod$ ($\tauBG$, $\tauFG$,$\tauA$, $\tauB$)
are optimized through a Bayesian optimization process~\cite{web-BayesianOpt}
on the entire dataset for each $X$ with the overall $\fscore$ score
as optimization criterion. Let us note that the case $X=1$ corresponds
to $\SemanticBackgroundSubtraction$ as semantic information is available
for all frames. To show the importance of repeating the decision of
the semantic classifier only when relevant, we compare our algorithm
with two heuristics that can also extend $\SemanticBackgroundSubtraction$
in the case of missing semantic information. The first heuristic never
repeats the decision of the semantic classifier and the second heuristic
always repeats the decision of the semantic classifier without checking
if the pixel's value has changed. Note that the former corresponds
to $\ourMethod$ with $\tauA<0$ and $\tauB<0$ and the latter with
$\tauA$ and $\tauB$ chosen larger than the upper bound of color
distances.

\begin{figure}[t]
\begin{centering}
\includegraphics[width=1\columnwidth]{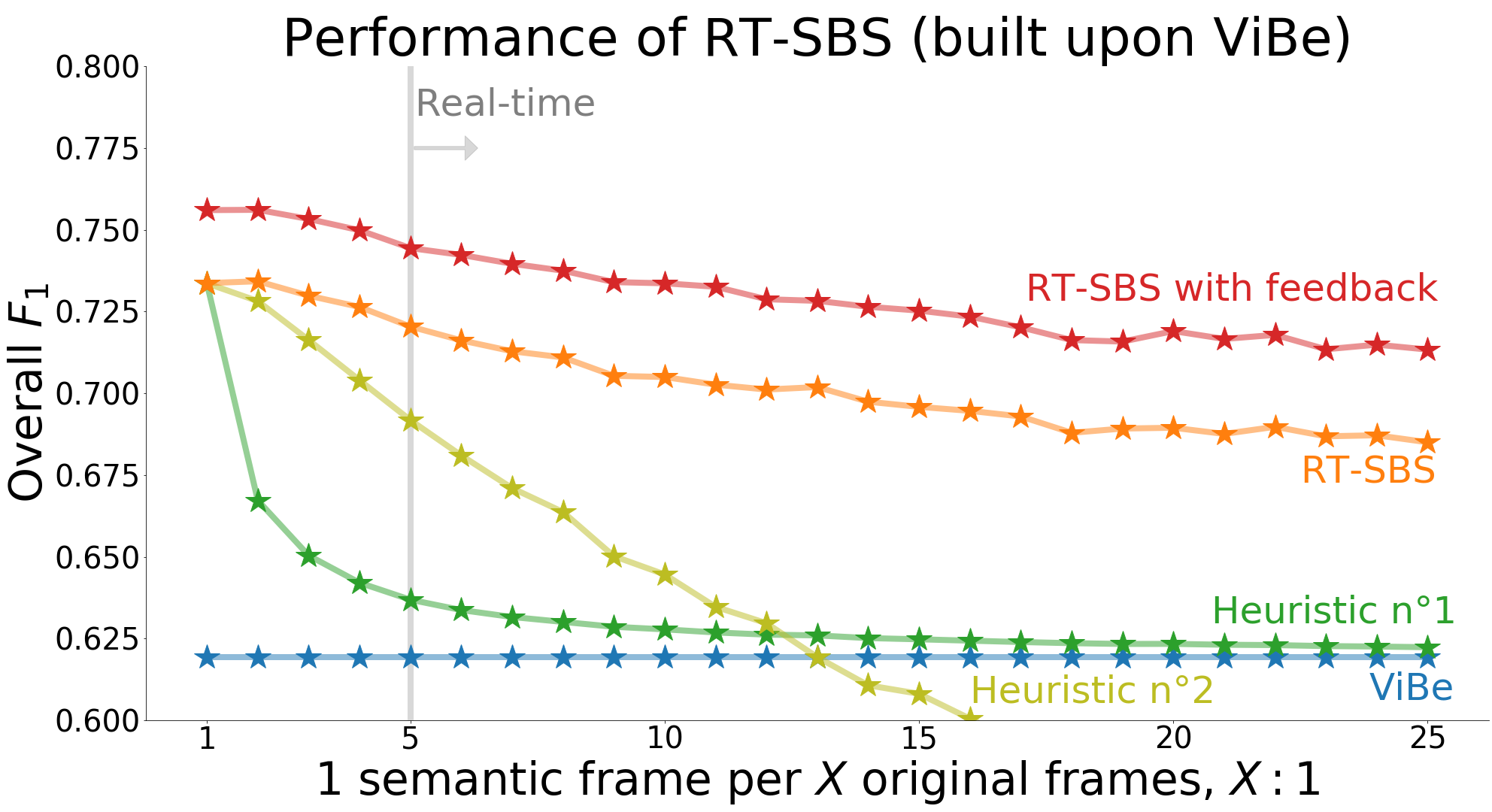}
\par\end{centering}
\caption{Overall $\protect\fscore$ scores obtained with $\protect\ourMethod$
(built upon ViBe) as a function of the semantic frame rate $\text{\ensuremath{\protect\samplingFactor X}}$
of \textcolor{tab_red}{$\protect\ourMethod$ with feedback}, \textcolor{tab_orange}{$\protect\ourMethod$
without feedback}, the \textcolor{tab_green}{ first heuristic},
the \textcolor{tab_olive}{ second heuristic}, and the \textcolor{tab_blue}{original
ViBe algorithm}.\label{fig:Overall-scores}}
\end{figure}

Figure~\ref{fig:Overall-scores} reports the performances of $\ourMethod$
built upon the ViBe $\text{\ensuremath{\BackgroundSubtraction}}$
algorithm. We draw five important observations from this graph: \textbf{(1)}
$\ourMethod$ always improves the performance compared to ViBe, even
when the semantic frame rate is low. \textbf{(2)} Its best real-time
performance (at $\text{\ensuremath{\samplingFactor 5}}$) is very
close to the one of $\SemanticBackgroundSubtraction$ (at $\text{\ensuremath{\samplingFactor 1}}$,
without feedback). \textbf{(3)} Both heuristics perform way worse
than $\ourMethod$, indicating that the semantic information should
only be repeated when relevant. This points out the importance of
the selectivity process introduced by the change detection algorithm
in $\ourMethod$. \textbf{(4)} Including a semantic feedback improves
the performance for all semantic frame rates. This is because the
internal model of the $\text{\ensuremath{\BackgroundSubtraction}}$
algorithm is improved, and thus its decisions are better overall.
Even at $\samplingFactor 5$, our algorithm with feedback surpasses
the performance of $\SemanticBackgroundSubtraction$. \textbf{(5)}
The best real-time performance is achieved with the feedback at $\samplingFactor 5$
with a $\fscore$ score of $0.746$. We compare this score with the
top-5 state-of-the-art unsupervised background subtraction algorithms
in Table~\ref{tab:comparison}. As can be seen, the performance of
our algorithm are comparable with the state-of-the-art algorithms
which are not real time. Furthermore, our algorithm performs better
than all real-time $\BackgroundSubtraction$ algorithms, making it
the state of the art for real-time unsupervised algorithms.

\begin{figure}[t]
\begin{centering}
\includegraphics[width=1\columnwidth]{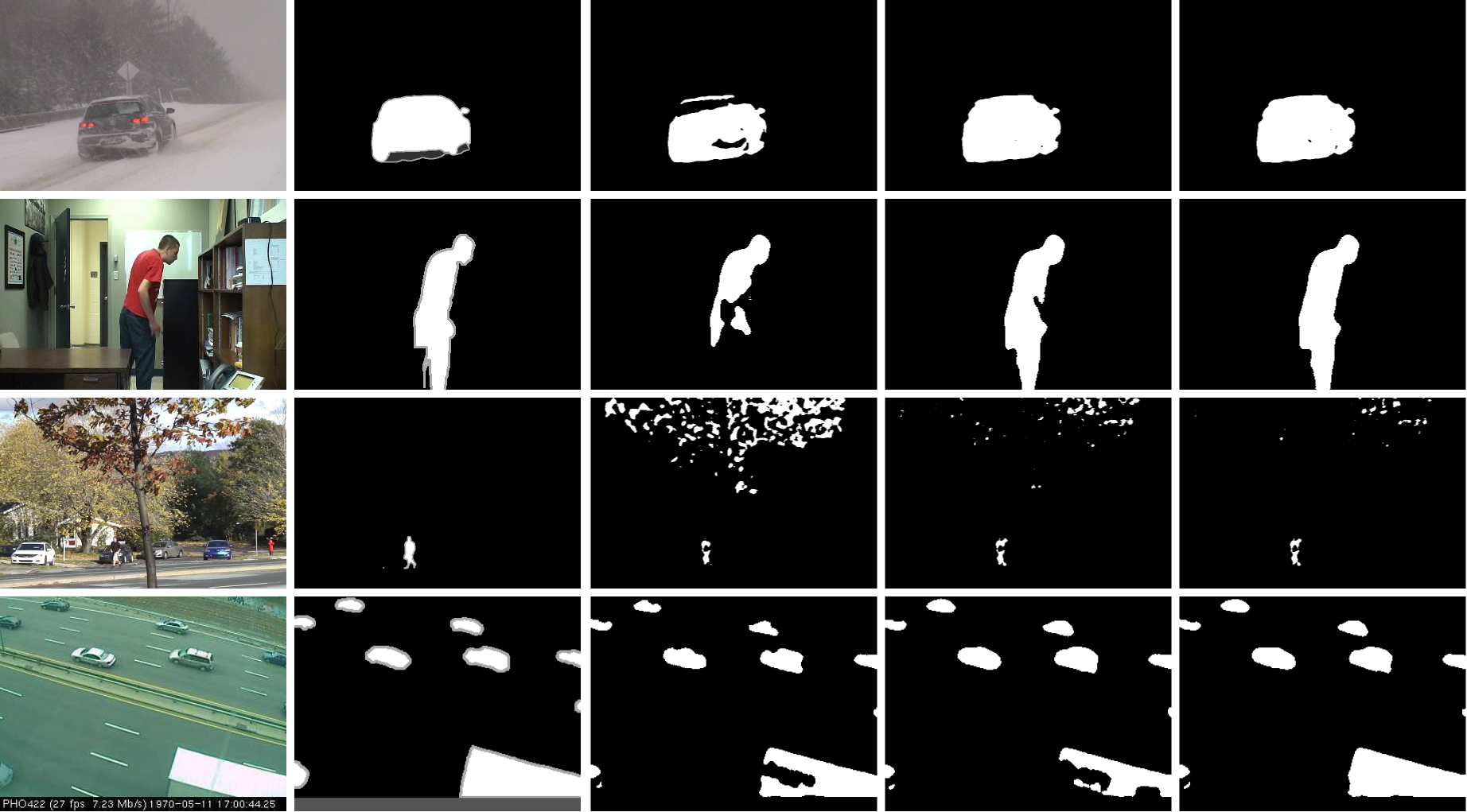}
\par\end{centering}
\caption{Qualitative evaluation of $\protect\ourMethod$. From left to right:
the original color image, the ground truth, the background subtraction
of ViBe, $\protect\ourMethod$, and $\protect\ourMethod$ with a feedback.\label{fig:results}}
\end{figure}

Also, we performed a scene-specific Bayesian optimization~\cite{web-BayesianOpt}
of the parameters; this leads to one set of parameters for each video.
It corresponds to a more practical use of a $\BackgroundSubtraction$
algorithm where its parameters are tuned for each application. With
this particular optimization, we obtain an overall $\fscore$ score
of $0.828$. This high score should not be compared with the others,
but still shows the great potential of our algorithm in real-world
applications. Finally, we display some results in Figure~\ref{fig:results}
showing our improvements qualitatively.

\begin{table}[t]
\centering{}{\footnotesize{}}%
\begin{tabular}{|c|c|c|}
\cline{2-3} 
\multicolumn{1}{c|}{Unsupervised $\BackgroundSubtraction$ algorithms} & $\fscore$ & fps\tabularnewline
\hline 
SemanticBGS ($\SemanticBackgroundSubtraction$ with IUTIS-5)~\cite{Braham2017Semantic} & $0.789$ & $\approx7$\tabularnewline
\hline 
IUTIS-5~\cite{Bianco2017Combination} & $0.772$ & $\approx10$\tabularnewline
\hline 
IUTIS-3~\cite{Bianco2017Combination} & $0.755$ & $\approx10$\tabularnewline
\hline 
WisenetMD~\cite{Lee2019WisenetMD} & $0.754$ & $\approx12$\tabularnewline
\hline 
WeSamBE~\cite{Jiang2018WeSamBE} & $0.745$ & $\approx2$\tabularnewline
\hline 
PAWCS~\cite{StCharles2015ASelfAdjusting} & $0.740$ & $\approx1-2$\tabularnewline
\hline 
ViBe~\cite{Barnich2011ViBe} & \multirow{1}{*}{$0.619$} & \multirow{1}{*}{$\approx152$}\tabularnewline
\hline 
\hline 
$\ourMethod$ at $X:5$ & $0.746$ & $25$\tabularnewline
\hline 
$\ourMethod$ at $X:10$ & $0.734$ & $50$\tabularnewline
\hline 
\hline 
$\ourMethod$ at $X:5$ & \multirow{2}{*}{\textbf{0.828}} & \multirow{2}{*}{\textbf{25}}\tabularnewline
 and scene-specific optimization &  & \tabularnewline
\hline 
\end{tabular}\caption{Comparison of the performance and speed of $\protect\ourMethod$ (build
upon ViBe + feedback) with the top-5 unsupervised $\protect\BackgroundSubtraction$
algorithms on the CDNet 2014 dataset and the previous best real-time
one. Our algorithm improves on some state-of-the-art algorithms while
being real time and surpasses all real-time $\protect\BackgroundSubtraction$
algorithms. The mean frame rates (fps) are taken from~\cite{Jodoin2014CDNetWebsite}.\label{tab:comparison}}
\end{table}

\section{Conclusion\label{sec:Conclusion}}

We presented a novel background subtraction algorithm, called Real-Time
Semantic Background Subtraction ($\ourMethod$), that extends the
semantic background subtraction ($\SemanticBackgroundSubtraction$)
algorithm for real-time applications. $\ourMethod$ leverages high-quality
semantic information which can be provided at any pace and independently
for each pixel and checks its relevance through time using a change
detection algorithm. We showed that our algorithm outperforms real-time
background subtraction algorithms and competes with the non-real-time
state-of-the-art ones. Python CPU and GPU codes of $\ourMethod$ are
available at \href{https://github.com/cioppaanthony/rt-sbs}{https://github.com/cioppaanthony/rt-sbs}.\newline\textbf{Acknowledgment.}
A. Cioppa has a grant funded by the FRIA, Belgium. This work is part
of a patent application (US 2019/0197696 A1). 

\bibliographystyle{IEEEbib}

\end{document}